%% file: main.tex
\documentclass{article}

\usepackage{arxiv}

\usepackage[utf8]{inputenc} % allow utf-8 input
\usepackage[T1]{fontenc}    % use 8-bit T1 fonts
\usepackage{hyperref}       % hyperlinks
\usepackage{url}            % simple URL typesetting
\usepackage{booktabs}       % professional-quality tables
\usepackage{amsfonts}       % blackboard math symbols
\usepackage{nicefrac}       % compact symbols for 1/2, etc.
\usepackage{microtype}      % microtypography
\usepackage{lipsum}		% Can be removed after putting your text content

\usepackage{xpatch}
\usepackage[backend=biber, style=numeric-comp, sorting=none, isbn=false, doi=true]{biblatex}

\DeclareSourcemap{
  \maps[datatype=bibtex]{
    \map{
      \step[fieldset=language, null]  % clear language
      \step[fieldset=year, null]
      \step[fieldset=month, null]
    }
  }
}

\DeclareFieldFormat[article]{title}{#1}  
\DeclareFieldFormat[inproceedings]{title}{#1}
\DeclareFieldFormat[misc]{title}{#1} 

\DeclareFieldFormat[article]{journaltitle}{\textit{#1}.}

\renewbibmacro{in:}{}
\DeclareFieldFormat[article]{volume}{\textbf{#1}}
\DeclareFieldFormat[article]{number}{(#1)}

\renewbibmacro*{journal+issuetitle}{%
  \usebibmacro{journal}%
  \setunit*{\addcomma\space}%
  \printfield{volume}%
  \setunit*{}%
  \printfield{number}%
}
\renewbibmacro*{note+pages}{%
  \printfield{note}%
  \setunit{\space}%
  \printfield{pages}%
  \newunit}
\renewbibmacro*{url+urldate}{%
  \ifentrytype{misc}{%
    \iffieldundef{doi}{% 没有 doi 才显示 url
      \iffieldundef{url}{}{%
        \printfield{url}%
      }%
    }{}%
  }{}%
}
\usepackage{xcolor}  % qq: colors
\usepackage{colortbl}
\usepackage{amsmath} % qq: math
\usepackage{mathrsfs} % qq: \mathscr{}
\usepackage{amsthm} % qq: for theorem indexing
\usepackage{doi} % qq: \doi{10.1234/abcd.5678}
\usepackage{graphicx}  % qq: figures
\usepackage{wrapfig} % qq: wrap figures
\usepackage{caption} % qq: caption*
\usepackage{subcaption} % qq: multi figure subcaptions
\usepackage{pifont} % qq: check, cross, star marks
\usepackage{multirow} % qq: multi-row in tables
\usepackage{longtable} % qq: cross-page table
\usepackage{booktabs} % qq: cross-page table
\usepackage{array} % vertical-align for tabular
\usepackage{makecell} 
\usepackage{microtype}
\DisableLigatures{encoding = *, family = *}

\definecolor{rowgray}{RGB}{240,240,240}

\usepackage{varwidth}

\title{\textbf{A Pre-trained Reaction Embedding Descriptor Capturing Bond Transformation Patterns}}

\date{} 					% Or removing it

\author{
    Weiqi Liu\textsuperscript{1,2}, 
    Fenglei Cao\textsuperscript{2,*}, 
    Yuan Qi\textsuperscript{1,2,3,*},
    Li-Cheng Xu\textsuperscript{2,*, \href{https://orcid.org/0000-0002-9553-0412}{\includegraphics[scale=0.06]{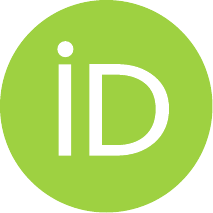}}}
}

\address{
    \textsuperscript{1} Artificial Intelligence Innovation and Incubation Institute, Fudan University, Shanghai, China \\
    \textsuperscript{2} Shanghai Academy of Artificial Intelligence for Science, Shanghai, China \\
    \textsuperscript{3} Zhongshan Hospital, Shanghai, China
}

\renewcommand{\headeright}{}
\renewcommand{\undertitle}{}

\begin{document}
\maketitle

\begingroup
\renewcommand{\thefootnote}{*}
\footnotetext{Corresponding authors. Email: xulicheng@sais.org.cn (L.-C.X.)}
\endgroup

\begin{abstract}
With the rise of data-driven reaction prediction models, effective reaction descriptors are crucial for bridging the gap between real-world chemistry and digital representations. However, general-purpose, reaction-wise descriptors remain scarce. This study introduces RXNEmb, a novel reaction-level descriptor derived from RXNGraphormer, a model pre-trained to distinguish real reactions from fictitious ones with erroneous bond changes, thereby learning intrinsic bond formation and cleavage patterns. We demonstrate its utility by data-driven re-clustering of the USPTO-50k dataset, yielding a classification that more directly reflects bond-change similarities than rule-based categories. Combined with dimensionality reduction, RXNEmb enables visualization of reaction space diversity. Furthermore, attention weight analysis reveals the model's focus on chemically critical sites, providing mechanistic insight. RXNEmb serves as a powerful, interpretable tool for reaction fingerprinting and analysis, paving the way for more data-centric approaches in reaction analysis and discovery.
\end{abstract}

% keywords can be removed

\input{chapters/ch1}

\input{chapters/ch2}

\input{chapters/ch3}

\input{chapters/ch4}

\input{chapters/ch5}

\printbibliography

\clearpage
\pagenumbering{arabic}
\appendix

\appendixtitle{\textbf{Supplementary Materials for ``A Pre-trained Reaction Embedding Descriptor Capturing Bond Transformation Patterns''}}
\makeappendixtitle

\input{chapters/si/s1}
\input{chapters/si/s2}
\end{document}

%% file: chapters/ch1.tex
\section{Introduction}

% % 半栏图
% \begin{wrapfigure}{r}{0.48\linewidth}
%   \centering
%   \vspace{-12pt}  
%   \includegraphics[width=0.92\linewidth, trim=0pt 0pt 0pt 0pt, clip]{figures/figure_1.pdf}
%   \captionof{figure}{\highlight{Types of chemical reaction descriptors.}}
%   \label{fig:descriptors0}
%   \vspace{-14pt}  
% \end{wrapfigure}

Precise organic synthesis serves as a cornerstone for creating high-performance functional molecules, including pharmaceuticals\cite{1_pharmaceuticals_1,1_pharmaceuticals_2}, advanced materials\cite{2_advances_materials_1,2_advances_materials_2}, and energy-related substances\cite{3_energy_related_substances_1,3_energy_related_substances_2}. In recent years, data-driven strategies for modeling reaction structure-performance relationships (SPRs) have gained significant momentum\cite{zuranski_predicting_2021,crawford_data_2021,zhang_bridging_2023,oliveira_when_2022}, offering powerful capabilities for predicting reaction reactivity\cite{ahneman_predicting_2018,yieldbert_Schwaller_2021,xu_highthroughput_2023,li_deep_2023,shi_prediction_2024,xiong_bridging_2025} and selectivity\cite{xu_towards_2021,guan_regio-selectivity_2021,qiu_selective_2022,reid_holistic_2019,caldeweyher_hybrid_2023,cuomo_feed-forward_2023}. These approaches are increasingly being applied to optimize and develop new organic reaction methodologies\cite{xu_transfer_2025,hou_machine_2024,gao_artificial_2024}. A critical step in bridging the physical world of chemistry with the in silico data-driven discovery is the encoding of chemical reactions into suitable representations\cite{gallegos_importance_2021,himanen_dscribe_2020}. Consequently, the design of effective and informative reaction encoding schemes remains a central challenge and a vibrant area of research. 

\begin{figure}[ht]
\centering
\includegraphics[width=0.5\textwidth]{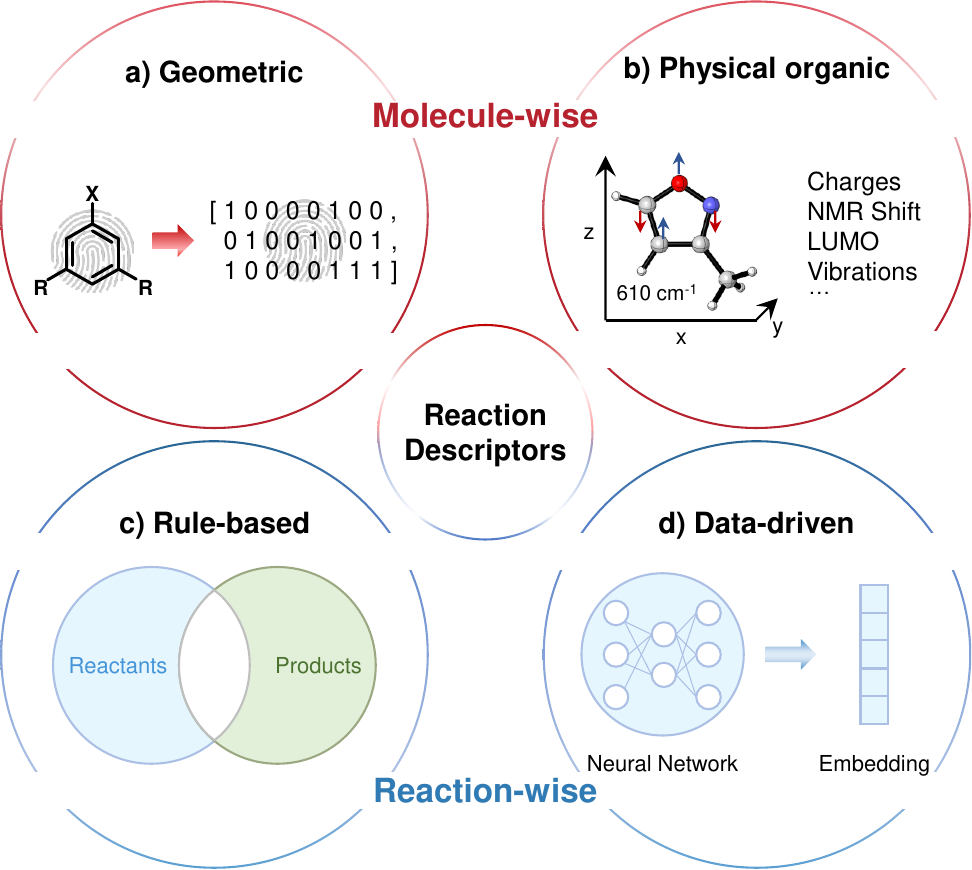}
\caption{Types of chemical reaction descriptors. \textbf{a)} Molecule-level descriptors based on 2D topological or 3D geometric structures. \textbf{b)} Molecule-level physical organic parameters derived from quantum chemical calculations. \textbf{c)} Reaction-level descriptors generated using predefined expert rules. \textbf{d)} Reaction-level descriptors generated by pre-trained deep learning models. }
\label{fig:figure1}
\end{figure}

Classical strategies typically compute steric or electronic descriptors for each molecular component within a reaction, which are then concatenated to form a holistic reaction representation\cite{xu_enantioselectivity_2023,li_reaction_2023}. These molecular descriptors can be broadly categorized into two classes: those derived solely from 2D topological or 3D structural information, such as multiple fingerprint features\cite{sandfort_structure-based_2020} (figure \ref{fig:figure1}a), Sterimol parameters\cite{verloop_development_1976}, average steric occupancy\cite{zahrt_prediction_2019}, among others \cite{hillier_combined_2003,behler_atom-centered_2011,bartok_representing_2013,xu_molecular_2021} and those requiring quantum chemical calculations to obtain more detailed physical organic features\cite{ahneman_predicting_2018,brethome_conformational_2019,singh_unified_2020,gallarati_reaction-based_2021} (figure \ref{fig:figure1}b). The latter has been exemplarily applied in the work of Doyle\cite{ahneman_predicting_2018,nielsen_deoxyfluorination_2018}, Sigman\cite{reid_holistic_2019,van_dijk_data_2023,haas_rapid_2025}, Grzybowski\cite{beker_prediction_2019,moskal_scaffolddirected_2021}, Reid\cite{plommer_extraction_2024}, Hong\cite{li_predicting_2020,yang_machine_2021,xu_enantioselectivity_2023}, and others\cite{hoque_deep_2022,chen_universal_2024,gao_machine_2024} to construct predictive models for reaction reactivity and selectivity.

However, constructing reaction encodings from molecular descriptors often faces an "alignment" challenge due to varying numbers of reaction components. A more direct approach is to generate reaction-wise descriptors based on the entire reaction\cite{schneider_development_2015,schwaller_mapping_2021,probst_reaction_2022}. A representative example is the differential reaction fingerprint (DRFP) developed by Reymond\cite{probst_reaction_2022} (figure \ref{fig:figure1}c), which computes the symmetric difference of circular substructures between reactants and products, hashing the result into a fixed-length binary vector to represent reaction changes. An alternative, fully data-driven strategy is exemplified by the rxnfp descriptor developed by Reymond and Schwaller\cite{schwaller_mapping_2021} (figure \ref{fig:figure1}d). This method employs a deep learning model trained for reaction classification, enabling its intermediate embeddings to intrinsically encode reaction information without relying on pre-defined expert rules. It is noteworthy, however, that the reaction class labels used for training this model were themselves generated by expert-defined rules.

In our recent work, we developed RXNGraphormer\cite{xu_unified_2025}, a pre-trained deep learning framework applicable to reaction performance prediction and synthesis planning. Through a specifically designed pre-training task that distinguishes real from fictitious reactions, the model learns the underlying patterns of bond formation and cleavage, rendering its internal embeddings—termed RXNEmb—a powerful general-purpose reaction encoding (figure \ref{fig:figure1}d). In this work, we explore the potential of RXNEmb as a reaction descriptor that inherently reflects bond transformation patterns. Specifically, we apply it to perform unsupervised clustering on reactions from standard datasets like USPTO-50k\cite{schneider_development_2015}, aiming to delineate reaction types in a data-driven manner. Additionally, we employ dimensionality reduction to visualize and compare the latent space distribution of RXNEmb encodings across diverse reaction datasets, contrasting the broad chemical space covered by general datasets with the focused regions occupied by specific reaction-type benchmark datasets. Furthermore, by visualizing the attention weights in the model's intermediate layers, we provide an interpretable analysis of the chemical bond change patterns learned by the model, shedding light on the underlying mechanisms it has learned.

%% file: chapters/ch2.tex
\section{Methods}

\subsection{Pre-training dataset}

The pre-training dataset for RXNGraphormer comprises two categories of reactions: approximately 6.8 million real organic reactions (represented as SMILES strings) sourced from open reaction databases, and an equal number of fictitious reactions constructed by randomly generating fictitious products via fragment exchange on the SMILES of real products. The total dataset encompasses over 13 million reactions. These fictitious reactions contain a multitude of bond transformation patterns that violate chemical rules. By performing contrastive learning on this large-scale dataset to distinguish between real and fictitious reactions, the model intrinsically learns the correct patterns of chemical bond changes.

\subsection{Reaction encoding and RXNEmb descriptor generation}
\label{sec:descriptor_generation}

\begin{figure}[ht]
\centering
\includegraphics[width=1\textwidth]{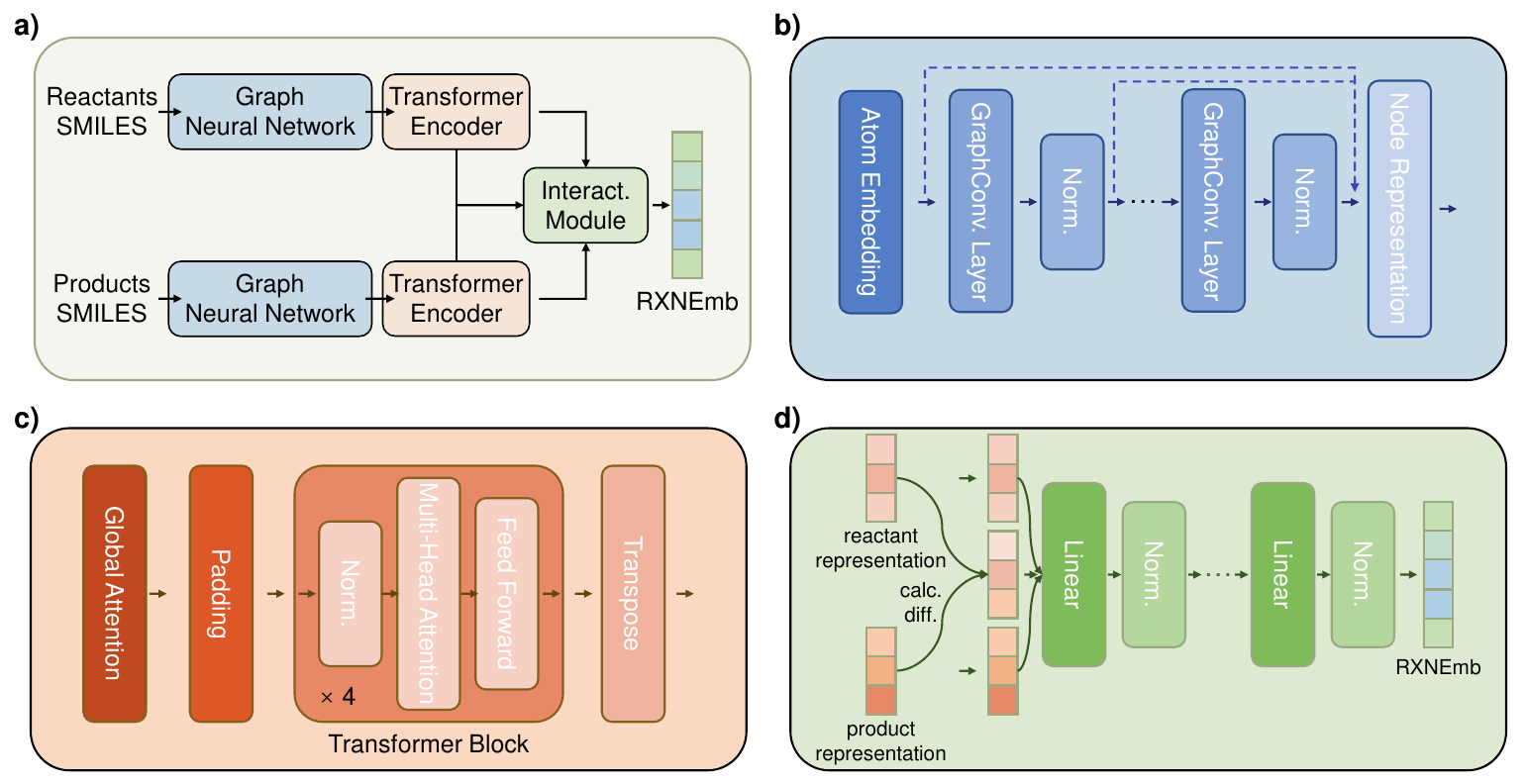}
\caption{Workflow for generating RXNEmb with RXNGraphormer. \textbf{a)} Schematic overview from reactant and product SMILES to RXNEmb. \textbf{b)} GNN module for intramolecular encoding. \textbf{c)} Transformer encoder module for intermolecular encoding. \textbf{d)} Interaction module for integrating reactant and product representations. Interact., interaction; Norm., normalization.}
\label{fig:figure2}
\end{figure}

We employ two parallel, structurally identical yet parameter-independent encoder sets to separately process the “reagent mixture” (labeled as “Reactants” in figure \ref{fig:figure2}a and consisting of reactants, solvents, and additives) and the products. Each encoder set contains a graph neural network (GNN)\cite{xia_mole-bert_2023} for encoding individual molecules and a Transformer module\cite{vaswani_attention_2017} for capturing intermolecular interactions.

The GNN for encoding a single molecule consists of 4 graph convolutional layers\cite{kipf_semi-supervised_2017} (figure \ref{fig:figure2}b). Jumping Knowledge\cite{xu_representation_2018} connections are employed to aggregate information from multiple layers, mitigating over-smoothing and preserving fine-grained structural details from different depths. The output from the final layer is selected as the ultimate node representation. This configuration allows local atomic environment information to integrate with broader topological context. The generated node embeddings are then aggregated into a molecule-level representation via a global attention-based pooling layer, which assigns differential weights to atom nodes to emphasize chemically significant atoms or functional groups.

After encoding individual molecules, a padding layer aligns the sets of molecules of varying sizes, enabling the model to handle reactions with varying numbers of components. The aligned vector sequence is then processed by a 4-layer Transformer block, which utilizes multi-head self-attention to effectively capture interaction information among all molecules within the reaction system (figure \ref{fig:figure2}c).

After the reagent mixture and products are processed by their respective encoder sets, their vector representations are obtained. First, the difference between these two representations is computed, yielding a differential representation. Subsequently, this differential representation is concatenated with the original representations of the reactants and products, resulting in a comprehensive reaction encoding. This comprehensive encoding is then passed through an interaction module composed of linear and normalization layers, ultimately yielding a fixed-length reaction descriptor that captures global reaction information (figure \ref{fig:figure2}d). In the pre-training task of RXNGraphormer, this reaction encoding is fed into a downstream classifier for binary (real/fictitious) classification. After pre-training, the reaction encoding generated by the model is defined as the proposed reaction descriptor, RXNEmb.

%% file: chapters/ch3.tex
\section{Results}

\subsection{Data-driven reaction classification with RXNEmb}
\label{sec:recls}

\begin{figure}[ht]
\centering
\includegraphics[width=\textwidth]{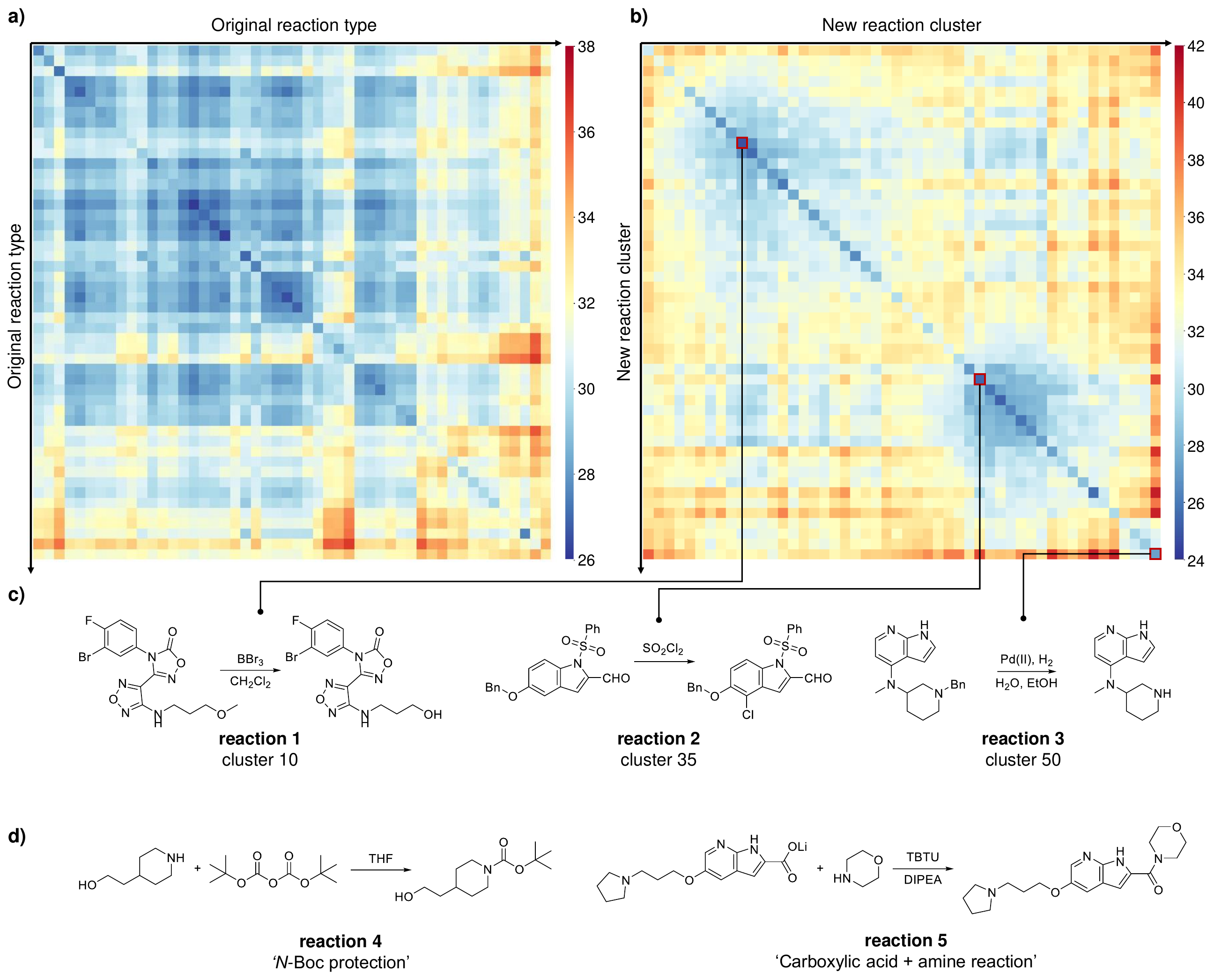}
\caption{Distance heatmaps and analysis of the USPTO-50k dataset in the RXNEmb latent space. \textbf{a)} Heatmap based on the original 50 reaction classes. \textbf{b)} Heatmap based on the newly defined 50 reaction clusters. \textbf{c)} Centroid reactions of three representative new reaction clusters. \textbf{d)} Two reactions belonging to the same new cluster (cluster 10) but previously categorized under different original classes. THF, tetrahydrofuran; TBTU, \textit{O}-(Benzotriazol-1-yl)-\textit{N},\textit{N},\textit{N'},\textit{N'}-tetramethyluronium tetrafluoroborate; DIPEA, \textit{N},\textit{N}-Diisopropylethylamine.}
\label{fig:figure3}
\end{figure}

In our previous work\cite{xu_unified_2025}, we generated RXNEmb descriptors for the 50,000 reactions in the USPTO-50k dataset\cite{schneider_development_2015}, which were originally classified by the NameRxn software\cite{noauthor_nextmove_nodate} into 50 specific reaction types under 9 major categories based on expert-defined rules. Given that the formation and cleavage of chemical bonds are central to how chemists categorize reactions, we computed the pairwise distances between the RXNEmb descriptors of these 50 reaction types and visualized them as a distance heatmap (figure \ref{fig:figure3}a). In the heatmap, a bluer color indicates closer proximity in the model's latent space, while a redder color indicates greater distance. We observed that, in the vast majority of cases, reaction types considered identical or similar by chemical rules are indeed close in the latent space, while different types are farther apart. However, a few exceptions exist, likely due to the inherent subjectivity and non-uniformity of rule-based classification. Therefore, we attempted to categorize reactions directly based on the distances between their RXNEmb descriptors via clustering, to achieve a data-driven partitioning of the USPTO-50k dataset that more directly reflects the intrinsic patterns of bond changes.

We first used the Kennard-Stone (KS) algorithm\cite{kennard_computer_1969} to select 50 representative reactions that are distant from each other in the descriptor space as initial centroids. All remaining reactions were then assigned to their nearest centroid, effectively clustering the entire dataset into 50 new groups. The choice of 50 groups was made to align with the original number of classes; in practice, this number can be adjusted based on the desired granularity. Furthermore, optimal leaf ordering\cite{joseph_fast_2001} was applied to the clustering result, ensuring that categories close in the latent space also have adjacent indices in the sorted order.

We then computed the distances between these new 50 groups, resulting in a new heatmap (figure \ref{fig:figure3}b). A rough comparison between figures \ref{fig:figure3}a and \ref{fig:figure3}b reveals that after data-driven reclassification and sorting, the color (distance) transitions in figure \ref{fig:figure3}b are smoother and more continuous. This indicates that the new ordering of categories better reflects the true distribution of descriptors in the latent space. Notably, gradients are observed at both ends of the diagonal in the heatmap, where the color fades from blue at the center towards the edges. Meanwhile, the region at the bottom-right corner of the matrix appears distinctly separate. This pattern suggests that the reaction data in the latent space may naturally separate into three major groups with distinct bond-change patterns. We selected one representative cluster from each of these perceived groups (clusters 10, 35, and 50), which are highlighted with red boxes in the figure \ref{fig:figure3}b, and display their centroid reactions in figure \ref{fig:figure3}c. The representative reaction for cluster 10 (\textbf{reaction 1}) is a demethylation involving C–O bond cleavage and O–H bond formation; for cluster 35 (\textbf{reaction 2}), an aryl chlorination forming a new C–halogen bond; and for cluster 50 (\textbf{reaction 3}), an \textit{N}-benzyl deprotection involving C–N bond cleavage. Their distinct bond changes in the substrate lead to their separation into different clusters. It is noteworthy that \textbf{reactions 1} and \textbf{3}, while both classified under the broad expert-defined category of "Deprotection", involve different bond-change events, illustrating the diversity within that original category.

To further contrast the expert-rule-based and data-driven classifications, we analyzed two additional cases from USPTO-50k. The first case involves the original category 50, labeled "Methylation". The defining feature of this category is the formation of a new C–X single bond to introduce a methyl group, encompassing a wide range of bond types (e.g., C–N, C–O, C–C, C–S). Consequently, from a bond-change-centric perspective, reactions originally categorized as "Methylation" are distributed across nearly all of the newly formed clusters. The second case highlights reactions grouped within the same data-driven cluster but belonging to different original categories. As shown in figure \ref{fig:figure3}d, \textbf{reaction 4} (originally '\textit{N}-Boc protection') and \textbf{reaction 5} (originally 'Carboxylic acid + amine reaction') both involve the formation of a new amide bond and are therefore grouped into the same cluster (cluster 10) in our bond-change-based classification. Together, these examples underscore that classification with RXNEmb yields clusters grounded directly in the intrinsic data structure of bond transformations, offering an objective and consistent alternative to taxonomies based on expert-defined rules.

\subsection{Reaction space analysis and evaluation with RXNEmb}
\label{sec:react_space}

\begin{figure}[th]
\centering
\includegraphics[width=1\textwidth]{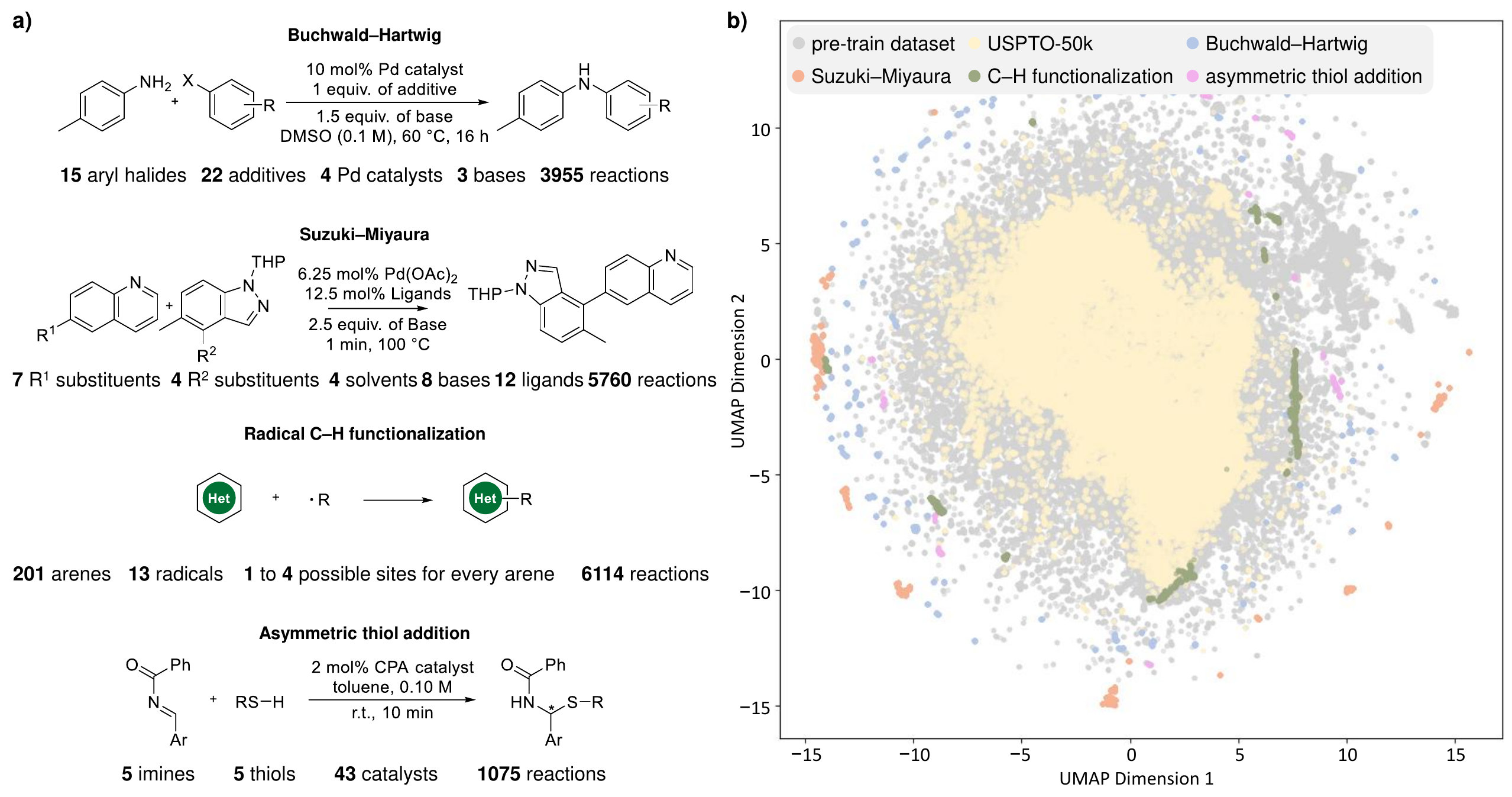}
\caption{Reaction space visualization of multiple reaction datasets. \textbf{a)} Four classic benchmark datasets for building SPR models. \textbf{b)} Reaction space visualization based on RXNEmb descriptors and UMAP. CPA, chiral phosphoric acid; r.t., room temperature.}
\label{fig:figure4}
\end{figure}

To gain deeper insights into the distribution characteristics of different reactions within the RXNEmb descriptor space, we generated RXNEmb descriptors for an additional set of datasets: the real reaction samples used for pre-training RXNGraphormer, and four classic benchmark datasets for SPR studies (figure \ref{fig:figure4}a), which are the Buchwald–Hartwig coupling\cite{ahneman_predicting_2018}, Suzuki–Miyaura coupling\cite{perera_platform_2018}, radical C–H functionalization\cite{li_predicting_2020}, and asymmetric thiol addition reaction datasets\cite{zahrt_prediction_2019}. We then employed UMAP\cite{mcinnes_umap_2020} to co-project the reaction features from all these datasets, along with USPTO-50k, into a two-dimensional space for visualization (figure \ref{fig:figure4}b). Detailed parameters for the UMAP fitting and transformation are provided in the Supplementary Materials.

Given the large scale of the pre-training dataset, 500,000 reactions were randomly sampled for plotting. This subset encompasses diverse chemical transformations, representing a broad reaction space. Figure \ref{fig:figure4}b reveals that data points from USPTO-50k are widely scattered across the 2D plane, reflecting its coverage of extensive reaction and bond-change types. In contrast, the four benchmark datasets, despite significant variations in substrates, catalysts, solvents, and other reaction conditions, are confined to very narrow regions of the reaction space. This is attributed to their inherently limited scope of chemical bond transformation types.

This analysis demonstrates that the combination of the RXNEmb reaction fingerprint and dimensionality reduction visualization provides a rapid assessment of the chemical diversity covered by a reaction dataset. It offers a valuable, intuitive reference for selecting or evaluating data when building SPR models for specific reaction types. By design, the pre-training learns bond-change patterns between substrates and products. However, in the benchmark datasets, reactivity and selectivity are driven largely by changes in catalysts and ligands—factors beyond the scope of simple bond transformations. Therefore, the pre-trained RXNEmb descriptor has inherent limitations in capturing these more subtle SPRs. Its primary utility lies in few-shot learning scenarios, where RXNEmb can serve as an effective fingerprint for measuring reaction similarity, enabling rapid retrieval of analogous reactions from large databases for data augmentation. Furthermore, although the pre-trained RXNEmb faces challenges in building high-precision SPR models directly, the fundamental knowledge of reactions learned during pre-training provides a foundational understanding. This knowledge can be effectively transferred, enabling the unified architecture of RXNGraphormer, after fine-tuning on specific tasks, to achieve accurate predictions of reactivity and selectivity\cite{xu_unified_2025}.

\subsection{Model interpretability via attention visualization}
\label{sec:attn}

Through reaction reclassification and dimensionality reduction visualization based on RXNEmb, our preceding analyses demonstrate that the pre-trained model effectively learns the patterns of bond formation and cleavage. Building on this, we further attempt to decipher how the deep learning model understands the process of chemical bond changes by interpreting the weight distributions within its self-attention modules.

\begin{figure}[ht]
\centering
\includegraphics[width=0.5\linewidth]{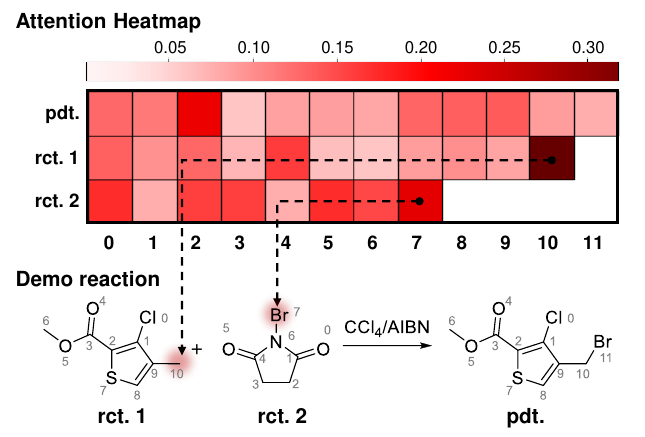}
\caption{Attention coefficient visualization for atoms in demo reaction. Color intensity reflects attention weight, with darker hues denoting higher values. rct., reactant; pdt., product.}
\label{fig:figure5}
\end{figure}

Taking the concrete example of a Wohl-Ziegler bromination reaction, we systematically analyzed the self-attention weights in the reactant and product encoders of RXNGraphormer. By tracing the attention distribution among atoms in the thiophene substrate (\textbf{rct. 1}), the brominating reagent \textit{N}-Bromosuccinimide (NBS, \textbf{rct. 2}) and the product (\textbf{pdt.}), and by computing and aggregating multi-layer attention scores, we constructed atom-level aggregated attention maps (figure \ref{fig:figure5}).

As shown in figure \ref{fig:figure5}, the carbon atom in the thiophene substrate that undergoes bromination receives the highest attention weight. Concurrently, the bromine atom in the NBS reagent, which serves as the bromine source, also attains the highest attention weight within its molecule. This indicates that during information processing, the model spontaneously assigns greater focus to the key sites (atoms or functional groups) where bond transformations occur. Therefore, the pre-trained RXNGraphormer not only provides the reaction-level descriptor RXNEmb but also, through analysis of its attention patterns, naturally highlights sites of mechanistic importance, offering insights akin to a chemist's reasoning about reaction processes.

%% file: chapters/ch4.tex
\section{Discussion}

This study introduces RXNEmb, a novel reaction-wise descriptor derived from the pre-trained deep learning model RXNGraphormer, capable of generating fixed-length fingerprints for organic synthesis reactions of variable components. As RXNGraphormer was pre-trained to distinguish real chemical reactions from fictitious ones containing erroneous bond changes, RXNEmb inherently encodes the ability to discern different bond formation and cleavage patterns. Leveraging this, we performed a re-clustering of the USPTO-50k dataset (containing 50 reaction types) using RXNEmb and the KS algorithm. This data-driven classification approach successfully groups into the same cluster reactions that were previously assigned to overlapping yet distinct rule-based categories, while also separating reactions within a single rule-based class that involve diverse bond-change scenarios.

Furthermore, RXNEmb, combined with dimensionality reduction techniques like UMAP, offers a method to explore the latent space of reactions. This facilitates an assessment of a dataset's diversity and similarity at the level of chemical transformations, provides an overview of data complexity for building SPR models, and enables the retrieval of similar reactions for data augmentation in few-shot learning contexts. Visualization of the model's self-attention weights reveals that the pre-trained model spontaneously focuses on atoms and functional groups critical to bond changes, assigning them higher weights, which directly elucidates the model's intrinsic capability to recognize different bond transformation types.

In conclusion, our work demonstrates the dual utility of RXNEmb in enabling data-driven reaction classification and providing mechanistic interpretation. With the growing availability of structured reaction data, RXNEmb can be applied to broader reaction datasets to enhance model interpretability and explore its transfer learning potential across different chemical domains. This contributes to paving the way for more automated, data-centric methodologies in chemical reaction analysis and discovery.

%% file: chapters/ch5.tex
\section{Acknowledgment}

Generous support by the National Natural Science Foundation of China (82394432 and 92249302, Y.Q.), and the Shanghai Municipal Science and Technology Major Project (Grant No. 2023SHZDZX02, Y.Q.). The model inference was conducted using the Inspire platform at SAIS.

\section{Data and code availability statement}

The code for generating RXNEmb and the experimental code for datasets such as USPTO-50k are available at \url{https://github.com/kzhoa/RXNEmb}.

\section{Conflict of interest}

The authors declare no competing conflict of interest.

\section{Author contributions}
L.-C.X. conceived the initial idea for the project. L.-C.X., F.C. and Y.Q. supervised the project. W.L. wrote the codes for reaction clustering, UMAP visualization, and extraction attention scores from RXNGraphormer. W.L. wrote the original draft. L.-C.X. reviewed and edited the manuscript.

%% file: chapters/si/s1.tex
\section{Supplementary Details on Visualizing Reaction Spaces and Attention Mechanisms}

In this section, we provide a comprehensive account of the analytical procedures that support the findings in Section~\ref{sec:react_space} (Reaction space analysis and evaluation with RXNEmb) and Section~\ref{sec:attn} (Model interpretability via attention visualization) of the main text. The purpose of these analyses is twofold: (1) to visually inspect how learned reaction representations differentiate among distinct reaction datasets, and (2) to interpret the internal processing of our model by examining its attention allocation over reactants and products.

\subsection{Details of Reaction Space Visualization}
\label{subsec:si_umap_detail}

Figure~\ref{fig:figure4} was constructed to reveal the latent geometry of reaction representations across five benchmark datasets. The procedure consists of the following steps:
\begin{enumerate}
\item \textbf{Descriptor Extraction}: For each reaction in the five baseline datasets, we computed the RXNEmb descriptor — a fixed-length vector encoding both reactant and product information based on a pretrained molecular embedding model. This yields a high-dimensional representation for every reaction instance.

\item \textbf{Dimensionality Reduction with UMAP}: We employed the Uniform Manifold Approximation and Projection (UMAP) algorithm to map the normalized descriptors into a 2D space for visualization. Key parameters were set as follows: \texttt{n\_neighbors=15}, \texttt{min\_dist=0.1}, \texttt{n\_components=2}, and Euclidean distance as the metric. These choices balance local neighborhood preservation with global structure visibility.

\item \textbf{Visualization}: The 2D coordinates output by UMAP were plotted using \texttt{matplotlib}. Each dataset was assigned a unique color/marker scheme to facilitate visual discrimination. A legend and axis labels were added for clarity.

\end{enumerate}

The resulting plot illustrates clusters corresponding to different datasets, reflecting both dataset-specific chemical biases and the discriminative power of RXNEmb representations. Notably, reactions from similar domains tend to form contiguous regions, while cross-dataset overlap remains limited, suggesting that the embeddings capture meaningful semantic distinctions.

\subsection{Details of Attetion Viasualization}
\label{subsec:si_attn_detail}

Figure~\ref{fig:figure5} visualizes the attention distribution learned by the global attention pooling module when processing reactants and products. The generation process is described below:

\begin{enumerate}

\item \textbf{Model and Layer Selection}: We used the trained transformer-based reaction prediction model described in Section~\ref{sec:descriptor_generation}. 
%The global attention pooling operates on token-level embeddings of SMILES strings for both reactants and products, producing a single context vector per molecule via weighted summation.
The global attention pooling operates on token-level embeddings (where each token represents an atom) of SMILES strings for both reactants and products, producing a single context vector per molecule via weighted summation.

\item \textbf{Attention Coefficient Extraction}: During a forward pass on a given reaction, we recorded the raw attention scores after softmax operator at the attention pooling layer. After applying softmax across atoms, we obtained a probability-like distribution (attention coefficients) indicating the relative importance of each atom.

\item \textbf{Color Mapping}: Atoms of the molecule were colored according to their attention coefficients. 
We employed a custom red-scale colormap ranging from light pink to deep red to encode attention magnitude, where lighter shades correspond to smaller attention coefficients and darker reds indicate larger values. This color-coding system provides an intuitive visual emphasis, directly highlighting the most influential atoms.

\item \textbf{Plotting}: The atoms of reactants and products were arranged side-by-side using \texttt{matplotlib}, with atoms belonging to the same reactant or product placed in the same row. Rows for reactants and products were positioned closely adjacent to one another. In addition, we have provided the expression for a demo reaction, and each atom is labeled with corresponding position index.

\end{enumerate}
This visualization reveals which substructures the model deems critical for predicting reaction outcomes. In many cases, functional groups or reactive centers receive disproportionately strong attention, aligning with domain knowledge. By comparing attention maps across datasets, we can also identify potential biases or generalization behaviors of the model.

%% file: chapters/si/s2.tex
\newpage
\section{Supplementary Details of RXNEmb-Based Reaction Reclassification}
\label{sec_si:reclassification_details}

This section provides detailed information on the 50 reclassified reaction clusters generated by the RXNEmb method, as mentioned in the main text in Section~\ref{sec:recls}. Each cluster is characterized by a representative "centroid reaction," which typically encapsulates the most common or representative reaction pattern within that cluster.
In Table \ref{tab:s1_recls_details}, we list all 50 clusters. For each cluster (Cluster \textit{i}, where i=1,2,…,50), the table contains the following three key pieces of information:

\begin{itemize}
\item \textbf{Cluster ID}: The unique identifier for the cluster (C1 to C50).
\item \textbf{Number of Reactions}: The total number of reactions assigned to that cluster. This value indicates the prevalence of that reaction pattern within the dataset.
\item \textbf{SMILES of Centroid Reaction}: The SMILES expression for the cluster's centroid reaction.
\end{itemize}

Collectively, these 50 clusters constitute a data-driven, mechanism-based partitioning of the analyzed reaction space. Examining the distribution of the number of reactions per group provides insight into the proportion of dominant reaction types in the dataset, while studying the SMILES of the centroid reactions allows for an intuitive understanding of the core chemical transformation for each cluster. This classification provides a structured foundation for subsequent reaction type analysis, model evaluation, or targeted dataset exploration.

\begingroup
% 设置行高（只影响当前表格）
\renewcommand{\arraystretch}{2}
\rowcolors{3}{rowgray}{white}  % 从第2行开始，偶数行灰色，奇数行白色
\begin{longtable}{
>{\centering\arraybackslash}m{1.2cm}
>{\centering\arraybackslash}m{1.5cm} 
>{\centering\arraybackslash\small}m{0.81\linewidth} 
}  % @{} 去掉两边间距
\caption{Details of reactions within each group}
\label{tab:s1_recls_details}\\

\toprule
\rowcolor{white}
{ClusterID} & {Num Reactions} & {\normalsize SMILES of Centroids} \\
\midrule
\endfirsthead  % 第一页表头

% 后续页也需要重新设置交替色

\toprule
\rowcolor{white}
{ClusterID} & { Num Reactions} & {\normalsize SMILES of Centroids} \\
\midrule
\endhead  % 后续页表头

\midrule
\rowcolor{white}
\multicolumn{3}{r@{}}{Continued on next page...} \\
\endfoot  % 每页底部（除最后一页）

\bottomrule
\rowcolor{white}
\multicolumn{3}{r@{}}{Table End.}
\endlastfoot  % 最后一页底部

% 表格内容
C1  & 365 & O=C(O)c1ccccc1.[Cl-]>>O=C(Cl)c1ccccc1 \\
C2  & 924 & \text{CCO.CN(c1ccnc2[nH]ccc12)C1CCCN(Cc2ccccc2)C1.Cl.O.[H][H].[OH-].[OH-].[Pd+2]}
\text{>>CN(c1ccnc2[nH]ccc12)C1CCCNC1.Cl.Cl.Cl} \\
C3  & 129 & CC(=O)O.CC(=O)OC(C)=O.O=C(O)c1cccc2c1OCCO2.O=[N+]([O-])O
>>O=C(O)c1cc([N+](=O)[O-])cc2c1OCCO2\\
C4  & 197 & \text{CO.COc1ccc(O)c(C)c1C.Cc1c(O)ccc(O)c1C.Cc1cc(O)ccc1O.[Zn]}
\text{\textnormal{>>}COc1c(C)cc(O)c(C)c1C} \\
C5  & 1104 & \text{O=Cc1cc2cc(OCc3ccccc3)ccc2n1S(=O)(=O)c1ccccc1.O=S(=O)(Cl)Cl}
\text{\textnormal{>>}O=Cc1cc2c(Cl)c(OCc3ccccc3)ccc2n1S(=O)(=O)c1ccccc1} \\
C6  & 103 & \text{CC1OC(CBr)OC1C.CCCOc1ccccc1N.O=C([O-])[O-].[Na+].[Na+]}
\text{\textnormal{>>}CCCOc1ccccc1NCC1OC(C)C(C)O1} \\
C7  & 86 & \text{CCCCCCCCCCO.COC(=O)OC.[Al].[Mg]\textnormal{>>}CCCCCCCCCCOC} \\
C8  & 1253 & \text{C1CCOC1.O.O=C(C1CC1)N1CCNCC1.[Al+3].[H-].[H-].[H-].[H-].[Li+].[Na+].[OH-]}
\text{\textnormal{>>}C1CN(CC2CC2)CCN1} \\
C9  & 86 & \text{CCCN(CCC)CCC.O=P(Cl)(Cl)Cl.Oc1nc2cc(F)ccc2n2cnnc12}
\text{\textnormal{>>}Fc1ccc2c(c1)nc(Cl)c1nncn12} \\
C10  & 96 & 
    \text{\small C1CCOC1.CC(C)(C)OC(=O)[C@@H]1CCC(=O)N1C(=O)OC(C)(C)C.CI.C[Si](C)(C)[N-][Si](C)(C)C}
    \text{\small .[Li+]\textnormal{>>}CC1C[C@@H](C(=O)OC(C)(C)C)N(C(=O)OC(C)(C)C)C1=O} \\
C11  & 312 &        
    \text{C1CCOC1.C1CCOC1.O.O=C(O)C(Br)c1ccccc1\textnormal{>>}OCC(Br)c1ccccc1} \\
C12  & 2929 & 
    \text{C=CCC(O)CC(O)c1cccc(Br)c1.ClCCl\textnormal{>>}C=CCC(O)CC(=O)c1cccc(Br)c1} \\
C13  & 88 & 
    \text{\small CC(C)(C\#N)N=NC(C)(C)C\#N.Cc1ccc(-c2ccccc2N([SH](=O)=O)C(C)(C)C)cc1.ClC(Cl)(Cl)Cl}
    \text{\small .O=C1CCC(=O)N1Br\textnormal{>>}CC(C)(C)N(c1ccccc1-c1ccc(CBr)cc1)[SH](=O)=O} \\
C14  & 299 & 
    \text{\small CN(C)C=O.CO.COC(=O)C12CCC(C(=O)O)(CC1)CC2.Cc1c(O)cccc1Br.ClCCl.ClCCl}
    \text{\small .O=C(Cl)C(=O)Cl\textnormal{>>}COC(=O)C12CCC(C(=O)Cl)(CC1)CC2} \\
C15  & 855 & 
    \text{CC(C)(C)[Si](C)(C)O[C@@H]1CO[C@H]2[C@@H]1OC[C@H]2O.ClCCl}
    \text{\textnormal{>>}CC(C)(C)[Si](C)(C)O[C@@H]1CO[C@@H]2C(=O)CO[C@@H]21}\\
C16  & 130 & 
    \text{O=S(=O)(O)O.O=S(=O)([O-])c1cccc(S(=O)(=O)[O-])c1.O=[N+]([O-])[O-].[Na+].[Na+].[Na+]}
    \text{\textnormal{>>}O=[N+]([O-])c1cc(S(=O)(=O)[O-])cc(S(=O)(=O)[O-])c1.[Na+].[Na+]} \\
C17  & 1475 & 
    \text{C1COCCO1.CCCC[Sn](CCCC)(CCCC)c1ccc(-c2cccs2)s1.Clc1ccc2cc3cc4ccccc4cc3cc2c1} 
    \text{.O=C(/C=C/c1ccccc1)/C=C/c1ccccc1.O=C(/C=C/c1ccccc1)/C=C/c1ccccc1}
    \text{.O=C(/C=C/c1ccccc1)/C=C/c1ccccc1.[Pd].[Pd]}
    \text{\textnormal{>>}c1csc(-c2ccc(-c3ccc4cc5cc6ccccc6cc5cc4c3)s2)c1} \\
C18  & 1230 & 
    \text{BrB(Br)Br.COCCCNc1nonc1-c1noc(=O)n1-c1ccc(F)c(Br)c1.ClCCl}
    \text{\textnormal{>>}O=c1onc(-c2nonc2NCCCO)n1-c1ccc(F)c(Br)c1} \\
C19  & 201 & 
    \text{CC\#N.CCCOc1nc(Cl)ccc1C(N)=O.NC(=O)c1ccc(Cl)nc1Cl.[H-].[Na+]\textnormal{>>}CC(C)COc1nc(Cl)ccc1C(N)=O}
    \text{} \\
C20  & 580  &\text{BrCc1ccccc1.CN(C)C=O.N\#CCc1ccc(O)cc1.N\#CCc1ccc(OCc2ccccc2)cc1.O.O=C([O-])[O-]}
    \text{.[I-].[K+].[K+].[K+]\textnormal{>>}CC(C\#N)c1ccc(OCc2ccccc2)cc1}\\
C21  & 107 & 
    \text{O=C(Cl)C(=O)Cl.O=[N+]([O-])C12C3C4C1C1C2C3C41[N+](=O)[O-]}
    \text{\textnormal{>>}O=[N+]([O-])C12C3C4C1C1(Cl)C2C3C41[N+](=O)[O-]}\\
C22  & 698 &   
    \text{C1CCOC1.C[SiH](C)OC(c1cccc(C\#N)c1)C(C)(C)C.[Al+3].[H-].[H-].[H-].[H-].[Li+]}
    \text{\textnormal{>>}C[SiH](C)OC(c1cccc(CN)c1)C(C)(C)C}\\
C23  & 371 &
    \text{O=Cc1cccc(Cl)c1Cl.O=S(=O)(O)O.O=[N+]([O-])[O-].[K+]}
    \text{\textnormal{>>}O=Cc1c([N+](=O)[O-])ccc(Cl)c1Cl}\\
C24  & 1148 &
    \text{CC\#N.CS(=O)(=O)c1ccc(C(=O)C2C(=O)CCCC2=O)c(Cl)c1.CS(=O)(=O)c1ccc(C(=O)O)c(Cl)c1}
    \text{.O=C([O-])[O-].O=C1CCCC(=O)C1.O=S(Cl)Cl}
    \text{.[Cl-].[K+].[K+].c1nc[nH]n1\textnormal{>>}CS(=O)(=O)c1ccc(C(=O)Cl)c(Cl)c1}\\
C25  & 447 &
    \text{CC(C)O.CCOCC.O=C(Cl)OCCl.c1ccncc1\textnormal{>>}CC(C)OC(=O)OCCl}
    \text{}\\
C26  & 2763 & 
    \text{C1CCOC1.CC(C)c1nnc2ccc(-c3cnco3)cn12.CN(C)C=O.C[Si](C)(C)[N-][Si](C)(C)C}
    \text{.O=C1CCC(=O)N1Br.[Li+]\textnormal{>>}CC(C)c1nnc2ccc(-c3ocnc3Br)cn12}\\
C27  & 188 &
    \text{CC(C)(C)C(C)(C)CO.ClCCl.O=[Cr](=O)([O-])Cl.c1cc[nH+]cc1\textnormal{>>}CC(C)(C)C(C)(C)C=O}
    \text{}\\
C28  & 169 &
    \text{CC=C(c1ccccc1)c1ccccc1.ClC(Cl)(Cl)Cl.O=C(OOC(=O)c1ccccc1)c1ccccc1.O=C1CCC(=O)N1Br}
    \text{\textnormal{>>}BrCC=C(c1ccccc1)c1ccccc1}\\
C29  & 304 &
    \text{CO.C[O-].Cl.O=C(CBr)c1ccc(O)cc1.[Na+]\textnormal{>>}COCC(=O)c1ccc(O)cc1}
    \text{}\\
C30  & 411 &
    \text{N\#CCc1csc2ccccc12.[Al+3].[Al+3].[Cl-].[Cl-].[Cl-].[H-].[H-].[H-].[H-].[Li+].[Na+].[OH-]}
    \text{\textnormal{>>}Cl.NCCc1csc2ccccc12}\\
C31  & 1695 &
    \text{C1CCOC1.CCN1CCC(C)(C)c2cc(C(C)C)cc(C=O)c21.C[Mg+].[Br-]}
    \text{\textnormal{>>}CCCN1CCC(C)(C)c2cc(C(C)C)cc(C=O)c21}\\
C32  & 753 &
    \text{CN(C)C=O.CN1C(=O)c2c(n[nH]c2Nc2ccccc2)N2C1=N[C@@H]1CCC[C@@H]12}
    \text{.Fc1ccc(-c2ccc(C(Br)Br)cc2)nc1.O=C([O-])[O-].[K+].[K+]}
    \text{\textnormal{>>}CN1C(=O)c2c(nn(C(Br)c3ccc(-c4ccc(F)cn4)cc3)c2Nc2ccccc2)N2C1=N[C@@H]1CCC[C@@H]12}\\
C33  & 212 &
    \text{C1CCOC1.CI.CN(CCC(=O)c1cccc(Cl)c1)C(=O)OC(C)(C)C.[H-].[Na+]}
    \text{\textnormal{>>}CC(CN(C)C(=O)OC(C)(C)C)C(=O)c1cccc(Cl)c1}\\
C34  & 204 &
    \text{CC(=O)NC1CCN(C)C(C)(C)C1C.CC(C)(C)c1cc(=C2CCCCC2)c(=C2CCCCC2)c2c1OC(=O)C2O}
    \text{.Cc1ccc(S(=O)(=O)O)cc1.ClCCl}
    \text{\textnormal{>>}CC(C)(C)c1cc(=C2CCCCC2)c(=C2CCCCC2)c2c1OC(=O)C2=O}\\
C35  & 695 &
    \text{CCO.COCc1ccccc1.OCCOCCOCCOCCOCCOCCOCCOCCOCCO.[Pd]}
    \text{\textnormal{>>}COCCOCCOCCOCCOCCOCCOCCOCCOCCO}\\
C36  & 9040 &
    \text{C1CCOC1.CCOC(C)=O.COC(=O)c1cccc2c1NC(=O)CO2.Cl.O=C([O-])[O-].[BH4-].[Na+].[Na+].[Na+]}
    \text{\textnormal{>>}COC(=O)c1cccc2c1NCCO2}\\
C37  & 1264 &
    \text{CC12CC3(C)CC(O)(C1)CC(NC(=O)OCc1ccccc1)(C2)C3.CN(C)C=O.O=C(Cl)CCCBr}
    \text{\textnormal{>>}CC12CC3(C)CC(NC(=O)OCc4ccccc4)(C1)CC(OC(=O)CCCBr)(C2)C3}\\
C38  & 327 &
    \text{Cc1c(Cl)cc(Cl)cc1Br.O=C(OOC(=O)c1ccccc1)c1ccccc1.O=C1CCC(=O)N1Br}
    \text{\textnormal{>>}Clc1cc(Cl)c(CBr)c(Br)c1}\\
C39  & 1807 &
    \text{COC(=O)Oc1ccc(Br)cc1C(C)(C)C.O=S(=O)(O)O.O=[N+]([O-])[O-].[K+]}
    \text{\textnormal{>>}COC(=O)Oc1cc([N+](=O)[O-])c(Br)cc1C(C)(C)C}\\
C40  & 1067 &
    \text{CC(=O)CCc1ccccc1.Cc1ncc(CO)c(C=O)c1O.O.O=P([O-])([O-])[O-].O=P([O-])([O-])[O-]}
    \text{.[K+].[K+].[K+]\textnormal{>>}CC(N)CCc1ccccc1}\\
C41  & 354 &
    \text{C1CCOC1.CNC(=O)CCCc1ccc(F)cc1.CO.[BH4-].[I].[Na+]}
    \text{\textnormal{>>}CNCCCCc1ccc(F)cc1}\\
C42  & 4432 & %这个表达式很复杂
    \text{C=C/C=C\textbackslash[C$@$H](C)[C$@$H](OC(N)=O)[C$@@$H](C)[C$@$H](O[Si](C)(C)C(C)(C)C)[C$@@$H](C)}
    \text{C(=O)N(C)C[C$@$H](C)[C$@@$H](O[Si](C)(C)C(C)(C)C)[C$@@$H](C)/C=C\textbackslash CO.ClCCl}
    % {C=C/C=C\[C@H](C)[C@H](OC(N)=O)[C@@H](C)[C@H](O[Si](C)(C)C(C)(C)C)[C@@H](C)C(=O)N(C)C[C@H](C)[C@@H](O[Si](C)(C)C(C)(C)C)[C@@H](C)/C=C\CO.ClCCl}
    \text{\textnormal{>>}C=C/C=C\textbackslash[C$@$H](C)[C$@$H](OC(N)=O)[C$@@$H](C)[C$@$H](O[Si](C)(C)C(C)(C)C)[C$@@$H](C)}
    \text{C(=O)N(C)C[C$@$H](C)[C$@@$H](O[Si](C)(C)C(C)(C)C)[C$@@$H](C)/C=C\textbackslash C=O}
    % {\textnormal{>>}C=C/C=C\[C@H](C)[C@H](OC(N)=O)[C@@H](C)[C@H](O[Si](C)(C)C(C)(C)C)[C@@H](C)C(=O)N(C)C[C@H](C)[C@@H](O[Si](C)(C)C(C)(C)C)[C@@H](C)/C=C\C=O}
    \\
C43  & 2379 &
    \text{C1=COCCC1.C1CCOC1.CO.Cl.ClC1CCCCO1.[H-].[Na+].c1c[nH]cn1\textnormal{>>}c1cn(C2CCCCO2)cn1}
    \text{}\\
C44  & 189 &
    \text{CCC=CCC=CCC=CCCCCCCCC(=O)O.ClC(Cl)Cl.O=C(Cl)C(=O)Cl}
    \text{\textnormal{>>}CCC=CCC=CCC=CCCCCCCCC(=O)Cl}\\
C45  & 2463 &
    \text{Nc1[nH]c(=O)[nH]c(=O)c1Br.Nc1ccccc1S.O=C([O-])[O-].OCCO.[K+].[K+]}
    \text{\textnormal{>>}Nc1ccccc1Sc1c(N)[nH]c(=O)[nH]c1=O}\\
C46  & 198 &
    \text{C[NH3+].Cc1c(C=O)cc(-c2ccccc2)n1S(=O)(=O)c1ccccc1.[BH3-]C\#N.[Cl-].[Na+]}
    \text{\textnormal{>>}CNCc1cc(-c2ccccc2)n(S(=O)(=O)c2ccccc2)c1C.Cl}\\
C47  & 68 &
    \text{CCN(CC)CC.C[Si]1(C)O[Si](C)(C)O[Si](C)(CCCO)O[Si](C)(C)O1}
    \text{.O=C(F)C(F)(OC(F)(F)C(F)(OC(F)(F)C(F)(OC(F)(F)C(F)(F)C(F)(F)F)C(F)(F)F)C(F)(F)F)C(F)(F)F.[F-]}
    \text{\textnormal{>>}C[Si]1(C)O[Si](C)(C)O[Si](C)}
    \text{(CCCOC(=O)C(F)(OC(F)(F)C(F)(OC(F)(F)C(F)(OC(F)(F)C(F)(F)C(F)(F)F)C(F)(F)F)C(F)(F)F)C(F)(F)F)}
    \text{O[Si](C)(C)O1}\\
C48  & 176 &
    \text{CC(C)Cc1ccccc1.ClBr.O=S=O.Oc1cccc(O)c1\textnormal{>>}CC(C)Cc1ccc(Br)cc1.CC(C)Cc1ccccc1}
    \text{}\\
C49  & 3490 &
    \text{C1CCOC1.CC(C)(C)C(=O)Nc1ccc(C(F)(F)F)cc1.CCCCCC.CI.O}
    \text{\textnormal{>>}Cc1cc(C(F)(F)F)ccc1NC(=O)C(C)(C)C}\\
C50  & 139 &
    \text{CC1COC(=O)O1.O=C1CCC(=O)N1Br.c1ccc2c(c1)Cc1ccccc1-2\textnormal{>>}Brc1ccc2c(c1)Cc1ccccc1-2}
    \text{} \\

\end{longtable}

\endgroup

\subsection{Original Reaction Classification of USPTO-50K}

To provide a baseline for comparison, this subsection details the original, manually-assigned reaction class labels of the USPTO-50K dataset, a widely used benchmark for chemical reaction prediction.

Expert annotators assigned these labels by inspecting atom-mapped reaction transformations and categorizing each according to a predefined taxonomy of reaction types (e.g., substitution, addition, elimination). Standardized guidelines were followed to ensure consistency, yet the process remains inherently subjective, especially for complex or multi-step reactions. Consequently, this manual classification results in a finite set of coarse-grained categories, which may not fully capture subtle mechanistic variations or overlapping reaction patterns.

Presenting these original labels serves two purposes: (1) to establish a reference point for evaluating data-driven reclassification methods (such as RXNEmb), and (2) to highlight the limitations of manual labeling for large-scale reaction datasets.

Table~\ref{tab:si_original_labels} enumerates all original reaction labels—including unique identifiers, major categorical groupings, and textual class names—thereby providing a comprehensive overview of the baseline taxonomy.

\begingroup
% 设置行高（只影响当前表格）
\renewcommand{\arraystretch}{2}
\rowcolors{3}{rowgray}{white}  % 偶数行灰色，奇数行白色
\begin{longtable}{
>{\centering\arraybackslash}m{1.1cm}
>{\centering\arraybackslash}m{1.5cm} 
>{\centering\arraybackslash}m{1.5cm} 
>{\centering\arraybackslash\small}m{0.4\linewidth} 
} 
\caption{Details of reactions within each group}
\label{tab:si_original_labels}\\

\toprule
\rowcolor{white}
{Index} & {Class ID} & {Major category}  & {\normalsize Class name} \\
\midrule
\endfirsthead  % 第一页表头

% 后续页也需要重新设置交替色

\toprule
\rowcolor{white}
{Index} & {Class ID} & {Major category}  & {\normalsize Class name} \\
\midrule
\endhead  % 后续页表头

\midrule
\rowcolor{white}
\multicolumn{4}{r@{}}{Continued on next page...} \\
\endfoot  % 每页底部（除最后一页）

\bottomrule
\rowcolor{white}
\multicolumn{4}{r@{}}{Table End.}
\endlastfoot  % 最后一页底部

% 表格内容
{1} & {1.2.1} & {1} & {Aldehyde reductive amination} \\
{2} & {1.2.4} & {1} & {Eschweiler-Clarke methylation} \\
{3} & {1.2.5} & {1} & {Ketone reductive amination} \\
{4} & {1.3.6} & {1} & {Bromo N-arylation} \\
{5} & {1.3.7} & {1} & {Chloro N-arylation} \\
{6} & {1.3.8} & {1} & {Fluoro N-arylation} \\
{7} & {1.6.2} & {1} & {Bromo N-alkylation} \\
{8} & {1.6.4} & {1} & {Chloro N-alkylation} \\
{9} & {1.6.8} & {1} & {Iodo N-alkylation} \\
{10} & {1.7.4} & {1} & {Hydroxy to methoxy} \\
{11} & {1.7.6} & {1} & {Methyl esterification} \\
{12} & {1.7.7} & {1} & {Mitsunobu aryl ether synthesis} \\
{13} & {1.7.9} & {1} & {Williamson ether synthesis} \\
{14} & {1.8.5} & {1} & {Thioether synthesis} \\
{15} & {2.1.1} & {2} & {Amide Schotten-Baumann} \\
{16} & {2.1.2} & {2} & {Carboxylic acid + amine reaction} \\
{17} & {2.1.7} & {2} & {N-acetylation} \\
{18} & {2.2.3} & {2} & {Sulfonamide Schotten-Baumann} \\
{19} & {2.3.1} & {2} & {Isocyanate + amine reaction} \\
{20} & {2.6.1} & {2} & {Ester Schotten-Baumann} \\
{21} & {2.6.3} & {2} & {Fischer-Speier esterification} \\
{22} & {2.7.2} & {2} & {Sulfonic ester Schotten-Baumann} \\
{23} & {3.1.1} & {3} & {Bromo Suzuki coupling} \\
{24} & {3.1.5} & {3} & {Bromo Suzuki-type coupling} \\
{25} & {3.1.6} & {3} & {Chloro Suzuki-type coupling} \\
{26} & {3.3.1} & {3} & {Sonogashira coupling} \\
{27} & {3.4.1} & {3} & {Stille reaction} \\
{28} & {5.1.1} & {4} & {N-Boc protection} \\
{29} & {6.1.1} & {5} & {N-Boc deprotection} \\
{30} & {6.1.3} & {5} & {N-Cbz deprotection} \\
{31} & {6.1.5} & {5} & {N-Bn deprotection} \\
{32} & {6.2.1} & {5} & {CO2H-Et deprotection} \\
{33} & {6.2.2} & {5} & {CO2H-Me deprotection} \\
{34} & {6.2.3} & {5} & {CO2H-tBu deprotection} \\
{35} & {6.3.1} & {5} & {O-Bn deprotection} \\
{36} & {6.3.7} & {5} & {Methoxy to hydroxy} \\
{37} & {7.1.1} & {6} & {Nitro to amino} \\
{38} & {7.2.1} & {6} & {Amide to amine reduction} \\
{39} & {7.3.1} & {6} & {Nitrile reduction} \\
{40} & {7.9.2} & {6} & {Carboxylic acid to alcohol reduction} \\
{41} & {8.1.4} & {7} & {Alcohol to aldehyde oxidation} \\
{42} & {8.1.5} & {7} & {Alcohol to ketone oxidation} \\
{43} & {8.2.1} & {7} & {Sulfanyl to sulfinyl} \\
{44} & {9.1.6} & {8} & {Hydroxy to chloro} \\
{45} & {9.3.1} & {8} & {Carboxylic acid to acid chloride} \\
{46} & {10.1.1} & {9} & {Bromination} \\
{47} & {10.1.2} & {9} & {Chlorination} \\
{48} & {10.1.5} & {9} & {Wohl-Ziegler bromination} \\
{49} & {10.2.1} & {9} & {Nitration} \\
{50} & {10.4.2} & {9} & {Methylation} \\

\end{longtable}

\endgroup